\documentclass[runningheads]{llncs}
\usepackage{graphicx}
\usepackage{longtable}
\usepackage{times}
\usepackage{graphicx}
\usepackage{amsmath}
\usepackage{amssymb}
\usepackage{array}
\usepackage{url}
\usepackage{cite}
\usepackage{enumitem}
\usepackage[linesnumbered,ruled,vlined]{algorithm2e}
\usepackage{soul,color}
\usepackage{booktabs} 

\usepackage{multirow}

\newcommand{\fdl}{federated learning\ }
\newcommand{\Fdl}{Federated learning\ }

\newcommand{\sync}{synchronize\ }

\newcommand{\etal}{\textit{et al.}}

\begin{document}
\title{Federated Generative Adversarial Learning}
%
%
%

\author{Chenyou Fan\inst{1} \and
Ping Liu\inst{2}
}

\institute{
Shenzhen Institute of Artificial Intelligence and Robotics for Society (AIRS), China \\ \email{fanchenyou@gmail.com} \and
Institute of High Performance Computing (IHPC), A*STAR, Singapore \email{pino.pingliu@gmail.com}
}

\maketitle              
\begin{abstract}
This work studies training generative adversarial networks under the federated learning setting. Generative adversarial networks (GANs) have achieved advancement in various real-world applications, such as image editing, style transfer, scene generations, etc. However, like other deep learning models, GANs are also suffering from data limitation problems in real cases. To boost the performance of GANs in target tasks, collecting images as many as possible from different sources becomes not only important but also essential. For example, to build a robust and accurate bio-metric verification system, huge amounts of images might be collected from surveillance cameras, and/or uploaded from cellphones by users accepting agreements. In an ideal case, utilize all those data uploaded from public and private devices for model training is straightforward. Unfortunately, in the real scenarios, this is hard due to a few reasons. At first, some data face the serious concern of leakage, and therefore it is prohibitive to upload them to a third-party server for model training; at second, the images collected by different kinds of devices, probably have distinctive biases due to various factors,~\textit{e.g.}, collector preferences, geo-location differences, which is also known as ``domain shift". To handle those problems, we propose a novel generative learning scheme utilizing a federated learning framework. Following the configuration of federated learning, we conduct model training and aggregation on one center and a group of clients. Specifically, our method learns the distributed generative models in clients, while the models trained in each client are fused into one unified and versatile model in the center. To the best of our knowledge, this is the first work on touching GAN training under a federated learning setting. We perform extensive experiments to compare different federation strategies, and empirically examine the effectiveness of federation under different levels of parallelism and data skewness.
\end{abstract}

\section{Introduction}

\begin{figure}
\begin{center}
\includegraphics[clip, trim=0 10 20 0, width=0.52\textwidth]{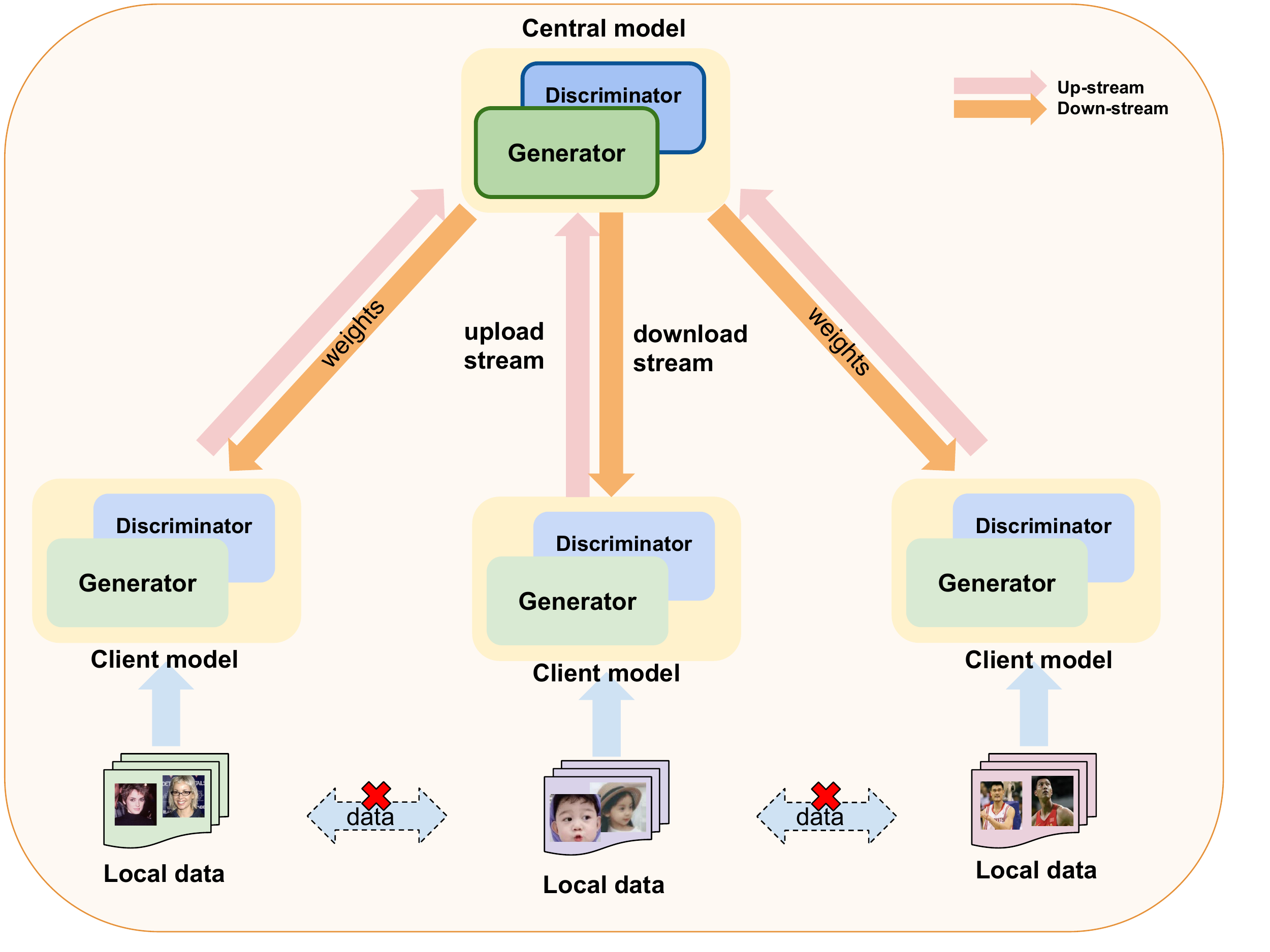}
\end{center}
\vspace{-10pt}
\caption{The task of generative learning under \fdl scheme. To preserve data privacy, remote devices exchange only model weights with a central server periodically to learn a global model. No data exchange would happen during any stage of communications.
}\label{fig:fed-gan}
\vspace{-10pt}
\end{figure}

Traditional machine learning methods require to gather training data into a central database and perform centralized training. However, as there are more and more edge devices such as smartphones, wearable devices, sensors, and cameras connecting to the World Wide Web, the data for training a model might be spread on various equipment. Due to privacy concerns, it might not be possible to upload all the data needed to a central node through public communications. How to safely access data on these heterogeneous devices to effectively train models has become an open research problem. To this end,
\fdl has become a rapidly developing topic in the research community~\cite{fedavg,zhao2018federated,li2019federated}, as it provides a new way of learning models over a collection of highly distributed devices while still preserving data privacy and communication efficiency.
\Fdl has witnessed many successful applications in distributed use cases such as smartphone keyboard input prediction~\cite{hard2018federated}, health monitoring~\cite{pantelopoulos2009survey,zhang2017zipml}, IoT~\cite{zhao2019mobile}, and blockchain~\cite{kim2019blockchained}.

Although federated learning has been applied successfully on discriminative model learning, how to apply it to generative learning is still under exploration. Generative Adversarial Network (GAN)~\cite{goodfellow2014generative} is one typical type of generative models which aims to gain generative capabilities based on game theory and deep learning techniques. Under the traditional machine learning framework, GANs have achieved huge successes in various applications such as realistic images/videos generation~\cite{goodfellow2014generative, vondrick2016generating, karras2019style,Yang_2019_ICCV}, face editing~\cite{he2019attgan,9110728}, domain adaptation~\cite{Luo_2019_ICCV,luo2020adversarial}, and style transferring~\cite{yang2018unsupervised}.
Though many efforts~\cite{fedavg, zhao2018federated, liu2018secure} have been made in boosting the performance of classification tasks with federated learning, there is little work of assessing whether existing \fdl framework works on generative learning or not. In real scenarios, we observe that data for generative learning are distributed among various equipment,~\textit{e.g.}, hand-written digits and signatures are stored in thousands of millions of mobile devices, facial images are stored in edge devices and IoT participants, etc. And therefore, it becomes urgent and necessary to understand whether the \fdl scheme is suitable for learning GANs.

In this paper, we propose a novel method of using a federated learning framework in GAN training; other than that, we discuss four strategies of synchronizing the local models and central model in the proposed method. We quantitatively evaluate the effectiveness of each strategy. Furthermore, we extensively study the GAN training quality under different data distribution scenarios and examine whether federated GAN training is robust to non-IID data distribution. In summary, our contributions include:
\begin{itemize}
    \item We formulate the federated generative adversarial learning outline with algorithm details, which is the first work in this direction to the best of our knowledge.
    \item We propose and compare four synchronization strategies for unifying local Generators and Discriminators to central models.
    \item We extensively study the training quality with different data distributions of different datasets under our framework.
\end{itemize}

\section{Related work}
Recently,
\fdl has become a rapidly developing topic in the research community~\cite{fedavg,zhao2018federated,li2019federated}, as it provides a new way of learning models over a collection of highly distributed devices while still preserving data privacy and communication efficiency.
\Fdl has witnessed many successful applications in distributed use cases such as smartphone keyboard input prediction~\cite{hard2018federated}, health monitoring~\cite{pantelopoulos2009survey,zhang2017zipml}, IoT~\cite{zhao2019mobile}, and blockchain~\cite{kim2019blockchained}.

Model averaging has been widely used in distributed machine learning ~\cite{zhang2015deep, chen2016revisiting, zhang2013communication}. In distributed settings, each client minimizes a shared learning target (in most cases, a loss function) on their local data, while the server aggregates clients' models by computing a uniform or bootstrap average of local model weights to produce the global model.
McMahan~\etal~\cite{fedavg} extended this learning strategy to the federated learning setting, in which data could be non-IID, and communications between clients and servers could be constrained. They proposed the FedAvg method to fuse local client models into a central model, and demonstrated its robustness of applying on deep learning models such as Convolutional Neural Networks (CNNs) and Recurrent Neural Networks (RNNs) with IID and moderately non-IID data. Zhao~\etal~\cite{zhao2018federated} discussed that FedAvg might suffer from weight divergence on highly skewed data, and several other works made attempts to propose robust federated learning in such cases~\cite{sattler2019robust,li2019convergence}. 
Recent work~\cite{liu2018secure} also discussed how to further improve the safety of communications during federated training using Additively Homomorphic Encryption~\cite{acar2018survey}.

Generative Adversarial Network (GAN)~\cite{goodfellow2014generative} aims to learn generative models based on game theory and deep learning techniques. Since its origin, GANs have witnessed huge successes in applications like generating realistic images~\cite{goodfellow2014generative, karras2019style, arjovsky2017wasserstein} and videos~\cite{vondrick2016generating} in computer vision areas. Conditional GAN (cGAN)~\cite{mirza2014conditional} is a natural extension of GAN which aims to generate images with given labels or attributes, such as human genders~\cite{radford2015unsupervised}, image categories~\cite{isola2017image} and image styles~\cite{karras2019style}.

To our best knowledge, the only similar work is from a technical report~\cite{tech_report19} which conceptually mentioned the possibility of using \fdl ideas in generative tasks. However, no further details were provided in this article.
Besides, a seemingly related work
\cite{hitaj2017deep} studied a type of adversarial attack under a collaborative learning environment. Their main purpose is to demonstrate that, by manipulating local training data, attackers could generate adversary training samples that harm the learning objective of a normal user. This is entirely different to the \fdl setting that no unsafe local data exchange should happen during client-client or client-server communications. 
\section{Approach}
\label{sec:approach}
We consider distributed GAN training on one center and a group of clients with the common communication-efficient \fdl framework~\cite{fedavg,konevcny2016federated}. Commonly, each client device possesses its local data with (usually) biased data distribution. E.g., personal devices are mostly used to take portraits, while surveillance cameras are often used to monitor street views. \textbf{We aim to train a unified central GAN model with the combined generative capacities of each client model.}  Yet we prohibit transferring any client data to the center as the communications between clients are costly and unsafe.  In the following sections, we will (1) investigate four types of synchronization strategies that arise naturally for federated GAN training, (2) briefly introduce the conditional GAN models' objective functions and architectures, and (3) summarize our proposed algorithm.

\subsection{Synchronization strategies}
\label{sec:sync_strategy}

FedAvg~\cite{fedavg} is a widely used \fdl framework that fuses client models to a central model by averaging the model weights. With FedAvg, the central model should be synchronized back (\textit{downloaded}) to clients periodically for training on local data, and after certain iterations, the clients \textit{upload} their local models to the central server for fusing into a new global model. 

In our study, however, training federated GANs is more complicated, as two parameter sets for generator (G) and discriminator (D) have to be communicated between the center and clients. This communication mechanism does not exist in a non-federated learning setting, while it is essential in a federated learning setting. How to guarantee an effective and efficient synchronization across the clients and the server becomes an open question.
We propose four types of synchronization strategies during communications. \textbf{Sync D\&G} synchronizes both the central model of D and G to each client. \textbf{Sync G} synchronizes central G model to each client. \textbf{Sync D} synchronizes central D model to each client. \textbf{Sync None} synchronizes neither G nor D from the center to clients. Please see Fig~\ref{fig:fed-gan} for the illustration of the overall process. These approaches still maintain the independence of each client during local training stage, while enable information propagation across clients during synchronization between server and clients' models.

\begin{figure}[h]
\begin{center}
\includegraphics[width=0.8\textwidth]{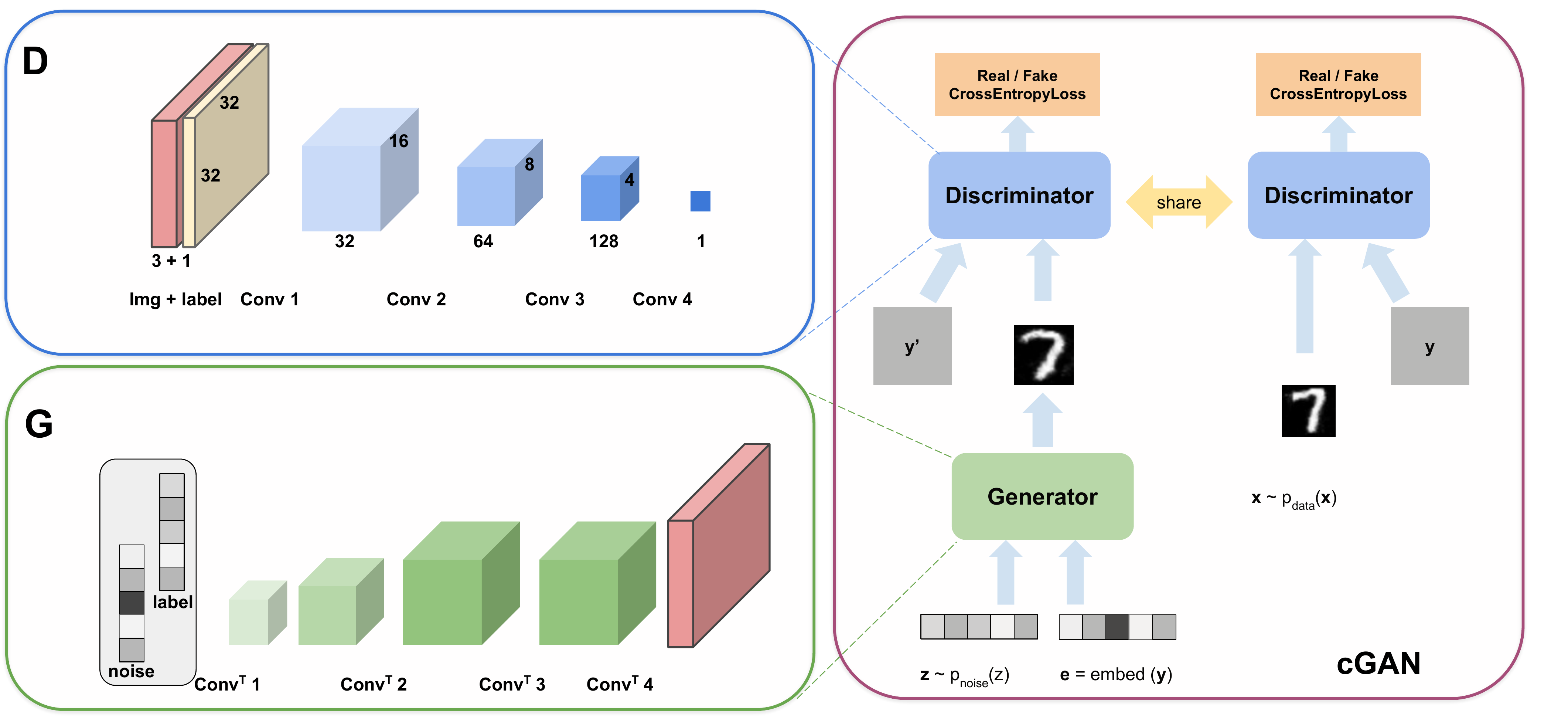}
\end{center}
\vspace{-10pt}
\caption{Architecture of a Conditional GAN with Generator G and Discriminator D. For G, the sampled noise vector $\mathbf{z}\sim \mathcal{N}(0,1)$ and a class label embedding vector $\mathbf{e}$ are fed into a deep neural network to generate a new image. For D, an image together with its label-spanned mask are fed into a deep neural network to predict whether the input image is real or fake.}\label{fig:ccgan}
\vspace{-10pt}
\end{figure}

\subsection{Conditional GAN (cGAN) model}
In our paper, one important mission is to analyze how data distribution affects GAN training in a federated setting. Therefore, we will study conditional GANs (cGANs)~\cite{mirza2014conditional}, which can manipulate the class distribution of generated images, e.g., generate "horse" images given the horse label. We will simulate different class distributions in clients' training data. Then we will
evaluate the training status of cGANs and analyze the robustness against skewed data distributions.
\begin{equation}
\min\limits_{G}\max\limits_{D} V(D,G) = \mathbb{E}_{\mathbf{x} \sim p_{data}} 
[\log D(\mathbf{x} | \mathbf{y})] + \mathbb{E}_{\mathbf{z} \sim p_{z}} 
[\log (1-D(G(\mathbf{z} | \mathbf{y})|\mathbf{y}))].
\label{eq:cgan}
\end{equation}
In a cGAN, the discriminator (D) and generator (G) play the minimax game with the objective function shown in Equation~\ref{eq:cgan}.  Intuitively, D learns to criticize the fidelity of given images while G learns to generate fake images with given labels as realistic as possible. 
In Figure.~\ref{fig:ccgan}, we demonstrate the typical architecture of D and G with Convolutional Neural Network structures. G takes a noise vector $\mathbf{z}$ and an additional label $\mathbf{y}$ to conditionally generate an image with a given label. For hand-written digit generation, the given labels indicate the digits from 0 to 9 to be generated. For the task of natural image generation, the provided labels indicate the image classes, e.g., on CIFAR-10, the classes are plane, car, bird, cat, deer, dog, frog, horse, ship, truck. D takes an image and its conditional label to predict whether it is real or fake. The conditional label is expanded to image size and attached along image channels. 

Our implementations of D and G networks follow DCGAN's~\cite{radford2015unsupervised} designs: D consists of four convolutional layers with BatchNorm~\cite{ioffe2015batch} and LeakyReLU~\cite{xu2015empirical}; G consists of four transposed convolutional layers with BatchNorm and LeakyReLU, followed by a $\tanh$ function to map features into normalized pixel values between -1 and 1. We alternatively update D by ascending its stochastic gradient
\begin{equation}
\label{eq:D}
\nabla_{\theta_D} \frac{1}{m} \sum_{i=1}^m \left [ \log D(\mathbf{x}_i | \mathbf{y}_i) + \log(1-D(G(\mathbf{z}_i|\mathbf{y}_i')|\mathbf{y}_i')) \right ]
\end{equation}
and update G by descending its stochastic gradient
\begin{equation}
\label{eq:G}
\nabla_{\theta_G} \frac{1}{m} \sum_{i=1}^m  \log(1-D(G(\mathbf{z}_i|\mathbf{y}_i')|\mathbf{y}_i'))
\end{equation}
in which $\mathbf{x}$ is a sampled batch of $m$ real images with true labels $\mathbf{y}$, $\mathbf{z}$ and $\mathbf{y'}$ are $m$ sampled noise vectors and labels. Intuitively, D is distinguishing real images from fake images conditioned on the given labels, while G is attempting to fool D by producing as realistic images as possible given designated labels. 

\subsection{Algorithm outline}
We summarize our algorithm of \textit{federated GAN learning} as follows. At each communication round, a subset of clients is randomly selected. Each client in the subset trains an updated model of GAN with Eq.~(\ref{eq:D}) and (\ref{eq:G}) based on their local data. After an epoch of training, the updated parameters of G and D are sent to the server via network communications. The server aggregates client models by weight averaging (or other model fusion techniques) to construct an improved central model. Finally, according to the chosen synchronization strategy mentioned in Section.~\ref{sec:sync_strategy}, each client pulls back the global model to reconstruct their local model. Each client then performs the next round of local model training. The above steps are repeated until convergence or some stopping criteria are met. The details of the algorithm are shown in Algorithm~\ref{alg:fed_GAN}. 

We implemented our algorithm and cGANs in PyTorch~\cite{paszke2017automatic}. The network parameters are updated by Adam solver~\cite{kingma2014adam} with batch size 64 and a fixed learning rate of $0.0002$. For each experimental setting, we train cGANs for at least 60 epochs with the federating step (communication between the center and clients) happening at the end of every epoch. We will release our source code for boosting further research.

\begin{algorithm}[htpb]
\DontPrintSemicolon
\KwIn{A global GAN model with parameters $(w_0^D,w_0^G)$ for Discriminator (D) and Generator (G) on central server $S$; local GAN models with parameters $\{(w_1^D,w_1^G),\ldots,(w_n^D,w_n^G)\}$ on $n$ clients $C=\{C_1,\ldots,C_n\}$; local private data $D=\{D_1,\dots,D_n\}$; Sync\_FLAG indicates whether to \sync central G and/or D back to clients.}
\KwOut{Fully trained global GAN model  $(w_0^D,w_0^G)$.}
 \For{\textup{communication round} $t = 1,2,\dots,T$ } {
    Select K random clients from all clients $C$\;
    \For{\textup{each client} $k = 1,2,\dots,K$ \textup{\textbf{in parallel}}}{ 
        Update discriminator $w^D_k$ of client k\;
        \textbullet~ sample a batch of $m$ real images $\mathbf{x}$ with true labels $\mathbf{y}$ \;
        \textbullet~ sample a batch of $m$ noise vectors $\mathbf{z}$ and labels $\mathbf{y'}$ from $D_k$\;
        \textbullet~ update $w^D_k$ by \textit{ascending} stochastic gradient in Eq~(\ref{eq:D}) \;
        \;
        Update generator $w^G_k$ of client k\;
        \textbullet~ sample a batch of $m$ noise vectors $\mathbf{z}$ and labels $\mathbf{y'}$ \;
        \textbullet~ update $w^G_k$ by \textit{descending} stochastic gradient in Eq~(\ref{eq:G}) \; 
  }
  Update central model by averaging client weights\;
  \ \ \ \ \ \ \ \ $w_0^D \gets \frac{1}{K} \sum_{k=1}^K w_k^D$\;
  \ \ \ \ \ \ \ \ $w_0^G \gets \frac{1}{K} \sum_{k=1}^K w_k^G$\;
  \If{Sync\_D\&G or Sync\_D}{
    \For{\textup{each client c in C} \textup{\textbf{in parallel}}}{
    $w_c^D \gets w_0^D$\;
    }
  }
  \If{Sync\_D\&G or Sync\_G}{
    \For{\textup{each client c in C} \textup{\textbf{in parallel}}}{ 
    $w_c^G \gets w_0^G$\;
    }
  }
 }
 \Return $(w_0^D,w_0^G)$
\caption{Federated Generative Learning algorithm.} \label{alg:fed_GAN}
\end{algorithm}





\section{Experiments}
\label{sec:exp}
We demonstrate the federated GAN training results on the MNIST and CIFAR-10 benchmark datasets. We first visualize samples of generated images for qualitative evaluation. Then we introduce the metrics to quantitatively evaluate GAN training results.
After that, we conduct experiments to evaluate the performance of different synchronization strategies proposed in Section~\ref{sec:sync_strategy}. By simulating IID and various non-IID data distributions, we further investigate the efficiency of model training with different data skewness levels. This enables us to probe the robustness of GAN training under \fdl framework.

\begin{figure}
\begin{center}
\includegraphics[clip, trim=0 10 0 0, width=1.0\textwidth]{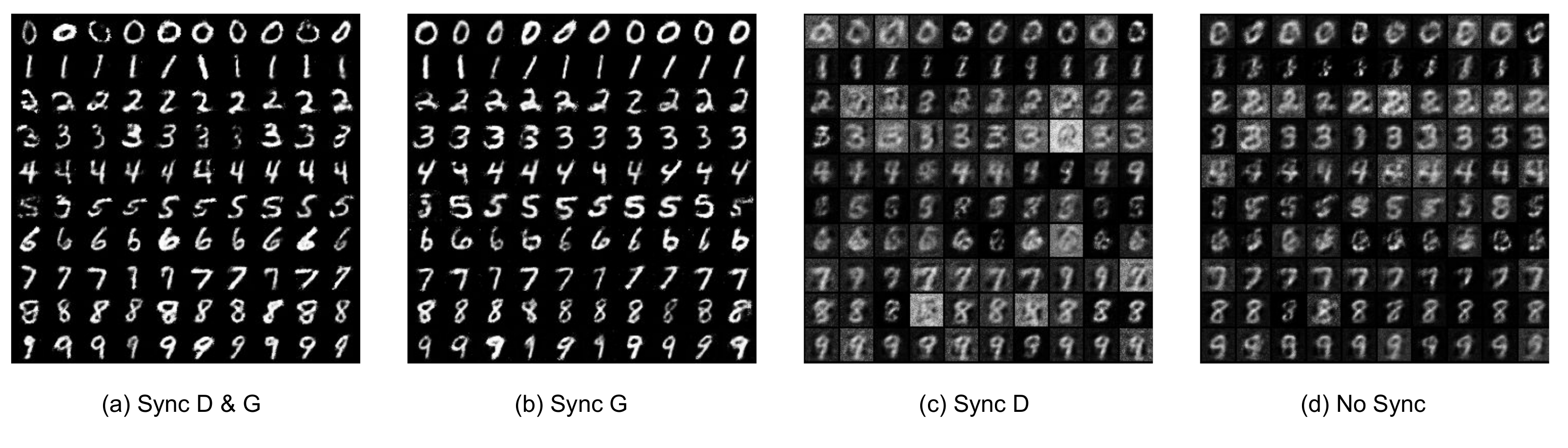}
\end{center}
\vspace{-10pt}
\caption{Samples of generated hand-written digits with different synchronization strategies. From (a) to (d), we (a) \sync both central D and G model to each client, (b) \sync only G model to each client while their individual D model is retained, (c) \sync only D model to each client while their G model is retained, (d) synchronize neither central D or G to any client.}\label{fig:sample_mnist}
\vspace{-15pt}
\end{figure} 

\subsection{Visualization}
\label{sec:visualization}
\begin{figure}
\begin{center}
\includegraphics[clip, trim=0 0 0 0, width=0.65\textwidth]{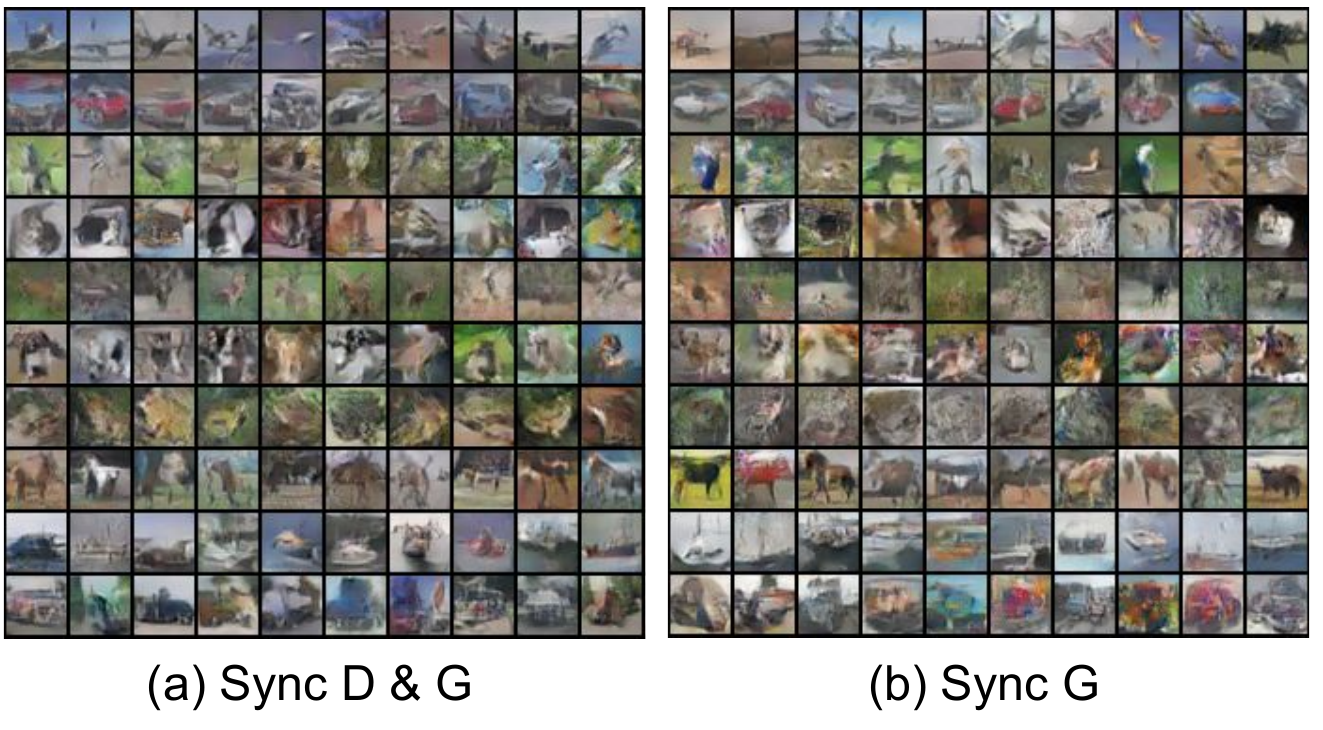}
\end{center}
\vspace{-10pt}
\caption{Generated images of CIFAR-10 classes - from top to bottom - plane, car, bird, cat, deer, dog, frog, horse, ship, truck. We show results of Sync D\&G and Sync G strategy.}\label{fig:sample_cifar}
\vspace{-10pt}
\end{figure}

In Fig.~\ref{fig:sample_mnist}, we show samples of digits generated by GANs trained with different synchronization strategies, which are illustrated in Section.~\ref{sec:sync_strategy}.  Specifically, in the federating step, a sampled collection of clients upload their Gs and Ds to the center. The center fuses their weights to form central model G and D with FedAvg or any other federated learning framework. As a quick reminder, strategy (a) Sync D \& G will synchronize both central D and G model to each client, (b) Sync G will synchronize central G model back to each client, (c) Sync D will synchronize central D model to each client, (d) synchronize neither central D or G to any client. Obviously, strategy (a) and (b) are visually better than (c) and (d), while (a) and (b) are comparable in image qualities. In Fig.~\ref{fig:sample_cifar}, we show samples of images generated for CIFAR-10 classes with strategy Sync D \& G and Sync G. We again found that these two strategies are comparable in visual quality. Curiously, the overall image quality is not as great as the generated digits in Fig.~\ref{fig:sample_mnist}. This is because of less training samples in CIFAR-10, and more complex patterns in natural images. However, how to improve GAN training with more data or with more capable neural network architecture is out of the scope of this paper. We will focus on how federated learning settings affect GAN training in the rest of the paper.

\subsection{Metrics}
We use two metrics for measuring the performance of cGANs on the image generation task.  
\textbf{1. Classification score (Score)} measures the ``reality" of a generator by using a pre-trained strong classifier $f_o$ to classify generated images.
In practice, we trained classifiers on MNIST and CIFAR-10, which yield a 99.6\% and 90\% accuracy on testing sets, respectively. We utilize the classifier as an oracle and apply it to generated samples to provide pseudo ground truth labels. 
Then we compare the pseudo ground truth labels with the conditional labels which are used to generate those images. The consensus between ground truth labels and conditional labels are taken as classification scores. Intuitively, the more realistic and fidelity the generated images are, the \textbf{higher} scores they will get.
\textbf{2. Earth Mover's Distance (EMD)}. Also known as Wasserstein distance~\cite{arjovsky2017wasserstein}, it measures the distance between the distribution of real data $P_r$ and generated data $P_g$. 
In practice, EMD is approximated by comparing average softmax scores of drawn samples from real data against the generated data such that
\begin{equation}
\begin{split}
EMD((\textbf{x}_r, \textbf{y}_r), (\textbf{x}_g,\textbf{y}_g)) = \frac{1}{N} \sum_{i=1}^N f_o(\textbf{x}_r^i)[y_r^i]-\frac{1}{N} \sum_{i=1}^N f_o(\textbf{x}_g^i)[y_g^i] 
\end{split}
\vspace{-5pt}
\end{equation}
in which $(\textbf{x}_r, \textbf{y}_r)$ are real data samples, $(\textbf{x}_g,\textbf{y}_g)$ are generated data samples, $f_o$ is the oracle classifier mentioned above. EMD measures a relative distance between real data and fake data. Obviously, a better generator should have a \textbf{lower} EMD by producing realistic images closer to real images.

\subsection{Result of different training strategies on IID data}
\begin{figure}
\begin{center}
\includegraphics[clip, trim=0 0 0 10, width=1.0\textwidth]{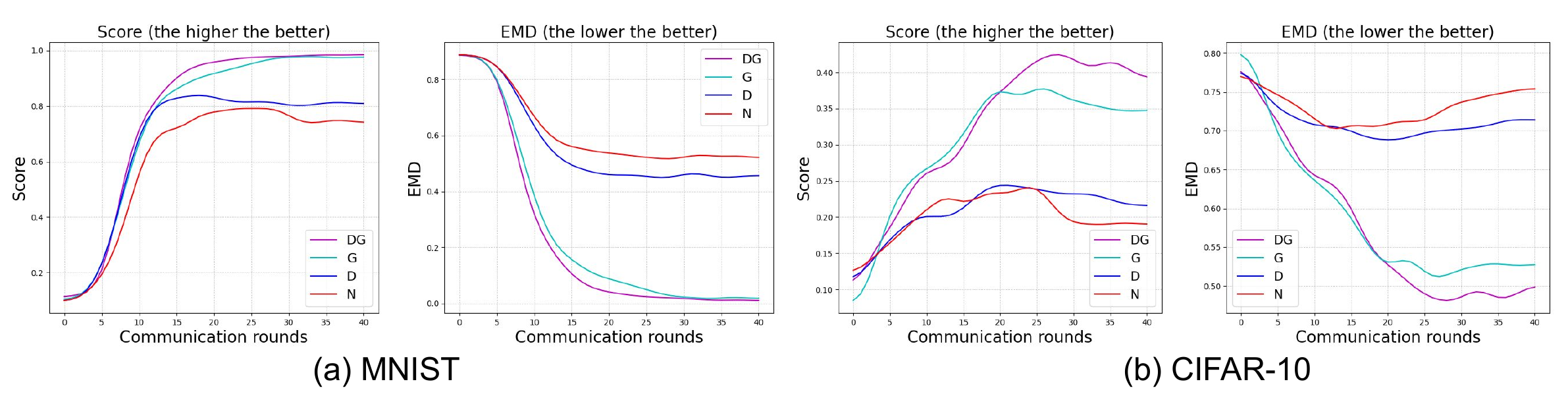}
\end{center}
\vspace{-10pt}
\caption{
Illustrations of training qualities influenced by different synchronization strategies. We show the Scores (the higher the better) and EMDs (the lower the better) on MNIST and CIFAR-10 datasets.}\label{fig:exp_tp2}
\vspace{-5pt}
\end{figure}

In an ideal case, data across the federated clients are independent and identically distributed (IID). We assume the IID condition and assume there are two federated clients. In Figure~\ref{fig:exp_tp2}(a), we show the training results of all four \sync strategies in two worker cases on the MNIST dataset. Congruent with visual intuitions from Fig.~\ref{fig:sample_mnist}, Sync D\&G (purple line), and Sync G (green line) are much better than Sync D (blue line) and Sync None (red line). The Scores and EMDs for the former two strategies are around 0.99 and 0.05. Scores for the latter two are about or lower than 0.8, while EMDs are above 0.4.

In Figure.~\ref{fig:exp_tp2}(b), we show training results on CIFAR-10 dataset and observe a similar trend: Sync D\&G significantly outperforms other methods. Sync G comes the second but is still much better than Sync D and Sync None. 
A question arises naturally: why is Sync D performing worse than Sync G? Our explanation is that by synchronizing central D across clients, the discriminative capacity of each client model grows rapidly. Unless we also synchronize G (as Sync D\&G does), the capacity of D exceeds G and thus rejects more samples of generated images. This harms and even stops the learning of G. Similar observation has also been reported by~\cite{arjovsky2017wasserstein, bang2018improved} in which they found the generator stops training if discriminator reaches optimum too early. Sync D \& G or Sync G would avoid this pitfall. In another aspect, by synchronizing G instead of both G and D, the communication costs could be reduced by about half in both upload and download streams. The trade-off between reducing communication costs and increasing training qualities should be considered case-by-case.
\textbf{For real-world applications when communication costs are essentially high, such as edge devices, we recommend to \sync G to reduce costs while sacrifice some generative capacity. Otherwise, we recommend to \sync both D and G}. In the following experiments, we will synchronize both D and G at default unless otherwise stated.

\subsection{Result of training GAN on different numbers of clients with IID data}
In this section, we investigate federated GAN training with IID data on different numbers of clients.  We build the training set of each client by randomly choosing 50\% of the total training samples with replacement to simulate IID data. We conduct three sets of experiments with $k=2,4,6$ federated clients. We also compare \fdl with a baseline method (k=1) by training GAN on a single client with the same amount of training data, simulating the situation that each client trains on its own data without federation. We show the results in Table.~\ref{tab:mnist_cifar_multi_worker_0.5} on both MNIST and CIFAR-10 dataset.


\begin{table}[htp]
\centering
\begin{tabular}{llccclccc}
\hline
\multirow{2}{*}{\shortstack[c]{Workers \\ Num k}} &  & \multicolumn{3}{c}{MNIST}                                                                                                                              &  & \multicolumn{3}{c}{CIFAR-10}                                                                                                                          \\ \cline{3-5} \cline{7-9} 
                                 &  & \begin{tabular}[c]{@{}c@{}}Optimal \\ Rounds\end{tabular} & \begin{tabular}[c]{@{}c@{}} ~ ~ Score ~ ~\end{tabular} & \multicolumn{1}{l}{EMD} &  & \begin{tabular}[c]{@{}c@{}}Optimal\\ Rounds\end{tabular} & \begin{tabular}[c]{@{}c@{}} ~ ~ Score ~ ~\end{tabular} & \multicolumn{1}{l}{EMD} \\ \cline{1-1} \cline{3-5} \cline{7-9} 
k = 1 (Local)   &  & 35  & 0.975 & 0.023 &  & 40 & 0.40 & 0.51                    \\ \hline
k = 2 (Fed)   &  & 25  & 0.990  & 0.004   &  & 25  & 0.428   & 0.475                    \\
k = 4 (Fed)                      &  & 25                                                         & \textbf{0.993}                                                       & \textbf{0.002}                   &  & 30                                                        & 0.432                                                            & 0.471                    \\
k = 6 (Fed)                      &  & 30                                                         & \textbf{0.994}                                                           & \textbf{0.002}                   &  & 35                                                        & \textbf{0.456}                                                            & \textbf{0.457}                    \\ \hline
\end{tabular}
\vspace{5pt}
\caption{Results of different numbers of federated workers on IID training data. The ``Optimal Rounds" column indicates how many communication rounds are needed for central models to reach optimal.
Best preforming numbers are highlighted for each column.}\label{tab:mnist_cifar_multi_worker_0.5}
\vspace{-15pt}
\end{table}

First, training GANs on federated clients (Fed) always outperforms training on a single worker (Local) with the same amount of local data. Moreover, we found that with increase in number of clients $k=2,4,6$, the metrics are slightly improving in terms of both Score and EMD on both MNIST (score: 0.99 v.s. 0.993 v.s. 0.994) and CIFAR-10 datasets (scores: 0.428 v.s. 0.432 v.s. 0.456). The higher evaluation score indicates that GAN training benefits from more federated workers, given IID training samples over clients. However, we also observed that training with more numbers of workers leads to slower convergence, as a trade-off for performance. On CIFAR-10, it took 25 communication rounds for central models to reach optimal when $k=2$, while it took 35 rounds when $k=6$.


\subsection{Result of training GAN with non-IID data}
Recent research~\cite{zhao2018federated, li2019convergence} observed that common \fdl methods such as FedAvg are not robust to non-IID data. In this section, we will verify the performance of GAN training with non-IID data of different data skewness levels. Let us suppose a dataset has $N$ classes, and there are $k$ clients. To simulate non-IID data across clients, we sort the data first by class. For each class, we randomly choose one client to allocate a fraction $p>0.5$ of the total training samples of that class, and then randomly allocate fraction $1-p$ of samples to other clients. This mimics a realistic scenario that data distribution is skewed across the clients, and the skewness is adjustable by $p$.
A larger $p$ indicates a higher degree of data skewness. We examine the training quality under different data skewness levels with different numbers of clients. 

\begin{table}[htp]
\centering
\begin{tabular}{llccclccc}
\hline
\multirow{2}{*}{\shortstack[c]{Workers \\ Num k}} &  & \multicolumn{3}{c}{CIFAR p=0.7}                                                                                                                              &  & \multicolumn{3}{c}{CIFAR p=0.9}                                                                                                                          \\ \cline{3-5} \cline{7-9} 
                                 &  & \begin{tabular}[c]{@{}c@{}}Optimal \\ Rounds\end{tabular} & \begin{tabular}[c]{@{}c@{}} ~ ~ Score ~ ~\end{tabular} & \multicolumn{1}{l}{EMD} &  & \begin{tabular}[c]{@{}c@{}}Optimal\\ Rounds\end{tabular} & \begin{tabular}[c]{@{}c@{}} ~ ~ Score ~ ~\end{tabular} & \multicolumn{1}{l}{EMD} \\ \cline{1-1} \cline{3-5} \cline{7-9} 

k = 2 (Fed)   &  & 30  & 0.40  & 0.50   &  & 30  & \textbf{0.37}   & \textbf{0.52}          \\
k = 4 (Fed)  &  & 35  & \textbf{0.44}  & \textbf{0.45}   &  & 40 & 0.35 & 0.57 \\
k = 6 (Fed)  &  & 40 & 0.42   & 0.48     &  & 30     & 0.31  & 0.58          \\ \hline
\end{tabular}
\vspace{5pt}
\caption{Results of training cGANs with non-IID data with data skewness level $p=0.7$ and $p=0.9$. We only show the results of CIFAR-10 due to page limits.}\label{tab:cifar_multi_worker_all}
\vspace{-15pt}
\end{table}

In Table~\ref{tab:cifar_multi_worker_all}, we demonstrate the experiment results on CIFAR-10 with $k=2,4,6$ and $p=0.7, 0.9$. Results for different $k$ are shown in different rows, and $p=0.7$ and $p=0.9$ are shown in separate columns. Obviously, the overall performance of $p=0.7$ is better than $p=0.9$, as the overall Score of $p=0.7$ is above $0.40$ while Score of $p=0.9$ is less than $0.37$. This indicates that the more skewed of the data distribution, the less effective federated training of GANs. We also found that a larger number $k$ of federated clients is more affected by skewed data distribution. For example, in Table~\ref{tab:mnist_cifar_multi_worker_0.5} IID case, as well as in Table~\ref{tab:cifar_multi_worker_all} (p=0.7), $k=6$ outperforms $k=2$ for both cases (IID and moderately non-IID). In contrast, in Table~\ref{tab:cifar_multi_worker_all} (p=0.9), we found that $k=6$ performs worse than $k=2$ in Score (0.30 v.s. 0.35, higher the better) and EMD (0.60 v.s. 0.55, lower the better) with highly non-IID(p=0.9) data. This accuracy drop can be explained by the weight divergence theory proposed by~\cite{zhao2018federated} such that more clients lead to faster divergence of model weights with non-IID training data.
We would like to encourage researchers to tackle the problem of \fdl of GANs with non-IID data in future.

\section{Conclusion}
We presented a comprehensive study of training GAN with different federation strategies, and found that synchronizing both discriminator and generator across the clients yield the best results in two different tasks. We also observed empirical results that federate learning is generally robust to the number of clients with IID and moderately non-IID training data. However, for highly skewed data distribution, the existing \fdl scheme such as \textit{FedAvg} is performing anomaly due to weight divergence.
Future work could further improve GAN training by studying more effective and robust model fusion methods, especially for highly skewed data distribution.

{\tiny
\bibliographystyle{splncs04}
\bibliography{prcv20,egbib}
}
\end{document}